\documentclass[journal]{IEEEtran}
\usepackage{amsmath,amsfonts}
\usepackage{algorithm}
\usepackage{array}
\usepackage{textcomp}
\usepackage{stfloats}
\usepackage{url}
\usepackage{verbatim}
\usepackage{graphicx}
\usepackage{cite}

\usepackage{comment}
\usepackage{amsmath,amssymb} 
\usepackage{times}
\usepackage{epsfig}

\usepackage{color}
\usepackage{multirow}
\usepackage{tabularx}
\usepackage{adjustbox}
\usepackage[font=footnotesize]{subcaption}
\usepackage{siunitx}
\usepackage[dvipsnames]{xcolor}
\usepackage [english]{babel}
\usepackage{setspace}
\usepackage{booktabs}
\usepackage{calc}
\usepackage{cite}
\usepackage{makecell}
\usepackage[table]{xcolor}

\usepackage{enumitem}
\usepackage{hyperref}
\hypersetup{
    pdfborder={0 0 0},     
    colorlinks=True       
}
\usepackage{pifont}  

\usepackage{algorithm}
\usepackage{xcolor}
\usepackage{algpseudocode}

\definecolor{darkgreen}{RGB}{0,127,0}
\definecolor{darkred}{RGB}{200,0,0}
\definecolor{pink}{RGB}{255, 50, 203} 
\definecolor{blue}{rgb}{0, 0, 1}

\definecolor{al_green}{RGB}{112,173,71}
\definecolor{al_blue}{RGB}{46,117,182}
\definecolor{al_red}{RGB}{197,90,17}

\def\greencheckmark{\textcolor{darkgreen}{\checkmark}}

\usepackage{xspace}
\newcommand{\eg}{e.g.\@\xspace}

\begin{document}

\title{\Large DeLTa: Demonstration and Language-Guided Novel Transparent Object Manipulation}
\author{{\footnotesize Taeyeop Lee$^{1*}$ \enspace Gyuree Kang$^{1*}$ \enspace Bowen Wen$^{2}$ \enspace Youngho Kim$^{1}$ \enspace Seunghyeok Back$^{3}$ \enspace In So Kweon$^{1}$ \enspace David Hyunchul Shim$^{1\dagger}$ \enspace Kuk-Jin Yoon$^{1\dagger}$}\\
{\footnotesize $^{1}$KAIST~~~~~~$^{2}$NVIDIA~~~~~~$^{3}$KIMM~~~~~~$^{*}$equal contribution~~~~~~$^{\dagger}$corresponding author}
}

\maketitle

\begin{abstract}
Despite the prevalence of transparent object interactions in human everyday life, transparent robotic manipulation research remains limited to short-horizon tasks and basic grasping capabilities.
Although some methods have partially addressed these issues, most 
of them have limitations in generalization to novel objects and are insufficient for precise long-horizon robot manipulation. To address this limitation, we propose DeLTa (Demonstration and Language-Guided Novel Transparent Object Manipulation), a novel framework that integrates depth estimation, 6D pose estimation, and vision-language planning for precise long-horizon manipulation of transparent objects guided by natural language task instructions. A key advantage of our method is its single-demonstration approach, which generalizes 6D trajectories to novel transparent objects without requiring category-level priors or additional training. 
Additionally, we present a task planner that refines the VLM-generated plan to account for the constraints of a single-arm, eye-in-hand robot for long-horizon object manipulation tasks.
Through comprehensive evaluation, we demonstrate that our method significantly outperforms existing transparent object manipulation approaches, particularly in long-horizon scenarios requiring precise manipulation capabilities.  
Project page: \url{https://sites.google.com/view/DeLTa25/}
\end{abstract}

\vspace{-10pt}

\section{INTRODUCTION}
Transparent objects are prevalent across real-world environments, including laboratories, kitchens, and manufacturing facilities. However, conventional depth sensors often fail to perceive these objects accurately. For example, commercial cameras~\cite{keselman2017intel, tadic2022perspectives} suffer from significant limitations when emitted infrared light undergoes refraction or reflection at transparent surfaces, producing erroneous or missing depth measurements. These sensor limitations cause substantial challenges for reliable robotic manipulation of transparent objects. Effective robotic manipulation in diverse scenarios requires both reliable perception capabilities and robust handling of various object types, with transparent objects being particularly challenging. While simple pick-and-place tasks may tolerate approximate 3D object locations~\cite{saxena2006robotic, fang2023anygrasp}, precise manipulation tasks demanding accurate grasping and placement require full 6D object pose estimation~\cite{wen2024foundationpose,tang2024automate,noseworthy2025forge,morgan2021vision}.

Transparent object pose estimation methods~\cite{qiu2025leveraging, zhang2024category, chen2022stereopose, liu2020keypose} often adopt category prior knowledge to estimate poses of novel object instances within the same category. As a result, robotic manipulation for transparent objects is inherently restricted to category-level object pose estimation~\cite{lee2023tta, jiang2304robotic, lee2021category}.  While category-level pose approaches have achieved promising results in generalizing to unknown objects within the same category, they struggle to generalize to novel objects beyond their trained categories. Moreover, their disregard of fine-grained object geometry limits applications to precise manipulation under certain task constraints (\eg align beverage bottles in a straight row when stocking a grocery shelf) even for novel object instances within the same trained category. This makes novel object instance pose estimation methods~\cite{wen2024foundationpose,labbe2022megapose,lee2025any6d} more desirable in such scenarios.

\begin{figure}[t]  
\centering  
\includegraphics[width=1.0\linewidth]{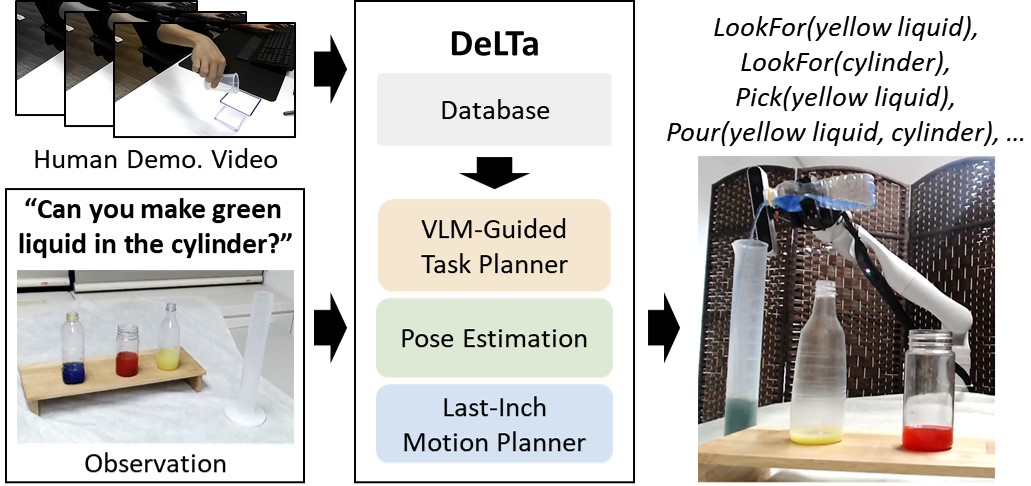}
\caption{DeLTa for Transparent Object Manipulation.}
\label{fig:teaser}
\vspace{-15pt}  
\end{figure}

\begin{figure*}[t]  
\centering  
\includegraphics[width=0.97\linewidth]{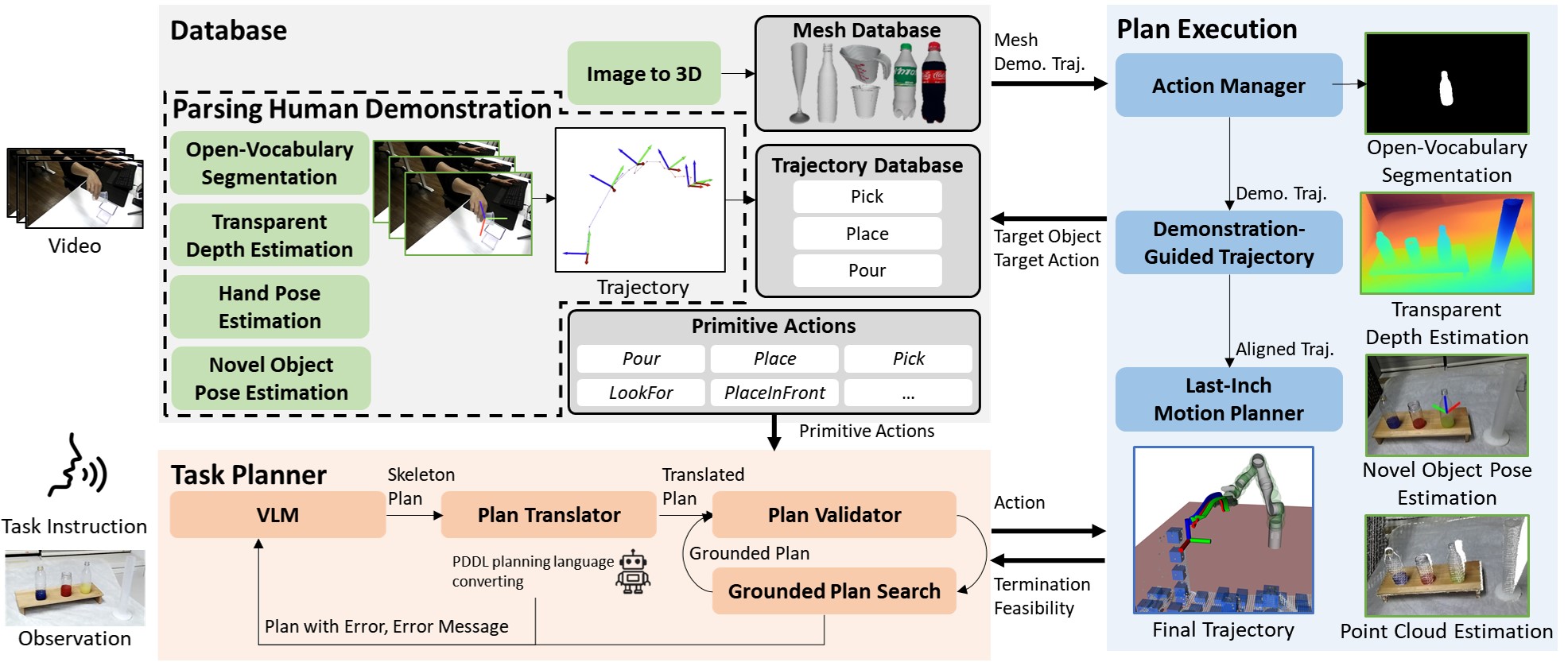}
\caption{Overview of our DeLTa framework.}
\label{fig:overview}
\vspace{-15pt}  
\end{figure*}

In terms of robotic manipulation policies for transparent objects, existing works~\cite{sajjan2020clear, shi2024asgrasp} have primarily focused on grasping diverse transparent objects.
Extending these methods to more diverse and challenging scenarios—such as target-constrained placement studied in this work—remains largely under-explored.
To address such complex tasks, recent advances in learning from demonstration have proven effective, offering a cost-efficient way to enable diverse trajectory actions~\cite{wen2022you,hsu2025spot,mandlekar2023mimicgen,garrett2024skillmimicgen}.
Compared to other demonstration data collection methods, such as robot demonstrations~\cite{vitiello2023one} or wearable devices~\cite{wang2024dexcap, chen2024arcap}, human demonstrations excel due to their intuitive operation and minimal hardware requirements.
Nevertheless, learning from human demonstrations has been mostly achieved on regular objects~\cite{wen2022you,hsu2025spot,lepert2025phantom,yin2025object}, leaving its extension to transparent objects insufficiently studied. 
This limitation arises because the complex optical properties of transparent objects pose challenges to visual sensing, which makes the extraction of action trajectories from demonstrations inherently difficult.
Consequently, insufficient capability for precise novel object pose estimation and the lack of diverse executable skills restrict long-horizon manipulation in real-world tasks.
Moreover, the limited exploration of language understanding (e.g., “Can you make a green liquid in the cylinder?”~\cite{burger2020mobile}) constrains progress toward natural-language-driven task execution—an essential step for human-robot interaction and generalizable manipulation.

To tackle these challenges, our contributions can be summarized as follows:
\begin{itemize}
\item{We propose DeLTa, to our best knowledge, the first framework that achieves precise and long-horizon manipulation tasks for transparent objects, guided by human video demonstration and language instructions, as illustrated in Fig.~\ref{fig:teaser}.} 
\item{For the first time, we explore 4D modeling of hand-object interaction information extracted from human demonstration video for transparent object manipulation, powered by recent advancements in stereo depth estimation, segmentation, and pose estimation.}
\item{We show that by combining pose-based trajectory retargeting with canonicalized object meshes, a single human demonstration per primitive skill can be transferred to novel objects, eliminating the need for per-object demonstrations.}

\item{We propose a VLM-guided planner that decomposes natural language instructions into multi-step actions, refines them with validation and search to enforce robot-specific constraints (e.g., single-arm, eye-in-hand), and retrieves object meshes and demonstration trajectories from pre-computed databases for transparent object manipulation.}

\end{itemize}

\section{Related Works}

\textbf{Transparent Object Manipulation: }
Most transparent object manipulation research has primarily focused on short-horizon grasping tasks, using either multiple views~\cite{ichnowski2021dex, duisterhof2024residual, jykim-2025-iccv, kerr2022evo}, single view~\cite{fang2022transcg, tang2024rftrans, jiang2022a4t, wang2025transdiff}. Most existing methods focus on reconstructing depth for grasping and have limited capabilities for long-horizon tasks that require precise manipulation from instructions.

\textbf{Transparent Object Pose Estimation: } 
One straightforward approach for achieving precise manipulation is through 6D object pose estimation, which suits robots operating in 3D space~\cite{wen2024foundationpose, wang2019densefusion, li2023fdct}. Most object pose estimation research~\cite{wen2024foundationpose, labbe2022megapose} has focused on non-transparent objects. While recent works~\cite{chen2022stereopose, zhang2024category, qiu2025leveraging} have addressed transparent object pose estimation, these methods remain limited to category-level understanding and still struggle to generalize to novel instances. Novel object pose estimation for transparent objects remains a challenging and open problem.

\textbf{From Human Demonstration to Robot Skills: } 
Teaching robots to perform human tasks necessitates intuitive and efficient ways that operate without requiring wearable sensors~\cite{wang2024dexcap, chen2024arcap} or teleoperation~\cite{vitiello2023one}. The most natural approach involves demonstrating a task once and enabling the robot to replicate the observed behavior~\cite{wen2022you, lum2025crossing}. However, these methods face significant limitations in transparent object perception and long-horizon manipulation scenarios, as they rely on traditional RGB-D perception and primarily focus on short-term tasks. Moreover, these demonstration-based methods generally lack obstacle-avoidance capability~\cite{wang2024cyberdemo}, a critical requirement for real-world deployment. 

\textbf{VLM-guided Long-Horizon Task Planning: } 
While significant progress has been made in applying VLMs to robotic long-horizon task planning~\cite{zhou2024isr, chen2024autotamp}, current approaches are often limited to simplified tasks or simulation environments and often assume privileged information of environmental objects~\cite{yang2025guiding, zhang2024fltrnn}, including ground-truth object poses. 
For instance, \cite{chen2024autotamp} repeatedly queries LLMs to generate symbolic plans but does not handle perception errors or real-time execution. 
This leaves a critical gap between noisy perception, motion planning, and symbolic planning, which occurs even more frequently for transparent object manipulation. Therefore, a unified framework for transparent object manipulation addressing these limitations is highly desirable.


\section{Method}
Our goal is to enable robots to execute manipulation tasks on novel transparent objects, given a task instruction, by leveraging single-object human demonstration trajectories.

\begin{itemize}[leftmargin=1em]
\item \textbf{Parsing Human Demonstration:} 
We obtain human demonstration trajectories (pick, place, pour) from single-object human demonstration videos. By leveraging foundation models for depth and pose estimation, we extract Cartesian-space trajectories of the object and store them in the trajectory database for the last-inch motion planner.
\item \textbf{Robot Action Execution:} 
Given task instructions and observations of novel transparent objects (different from the demonstration object), our VLM-guided planner generates a high-level task plan. The task plan is then converted into robot skills using demonstration trajectories and pose estimation during execution. 
The motion planner subsequently produces a collision-free path, refining it for precise and safe manipulation. 
\end{itemize}
We first explain our approach for parsing single-object demonstration trajectories from human videos (Section~\ref{subsec:human_demonstration}). We then describe the task planning process for long-horizon tasks from natural language instructions (Section~\ref{subsec:task_plan}). Finally, we detail how these trajectories are transferred to novel objects during robotic manipulation (Section~\ref{subsec:robot_execution}).

\subsection{Parsing Human Video Demonstration} \label{subsec:human_demonstration}
Fig.~\ref{fig:overview} (top-left) illustrates our human demonstration parsing pipeline, which extracts single-object demonstration trajectories from video demonstrations. In this work, we mainly consider three skill primitives: pick, place, and pour. 
For each skill primitive, we extract its trajectory from a single video demonstration of a randomly selected object. 
These extracted trajectories are then stored as trajectory database.
During robot execution, the action manager selects the trajectory-based skill corresponding to the primitive action, a basic task unit used by the task planner to compose high-level plans  (Sec.~\ref{subsec:robot_execution}).
Notably, our method requires only a single demonstration per skill, enabling cross-object transfer to manipulate novel objects.
This is in contrast to multiple separate trajectories for each target object, required as in prior works~\cite{hsu2025spot, lum2025crossing}. 

To build the trajectory databases, we require four key steps: stereo depth estimation, open vocabulary segmentation, novel object pose estimation, and hand pose estimation. We will explain each component in sequence.

\begin{figure}[t]  
\centering  
\includegraphics[width=0.9\linewidth]{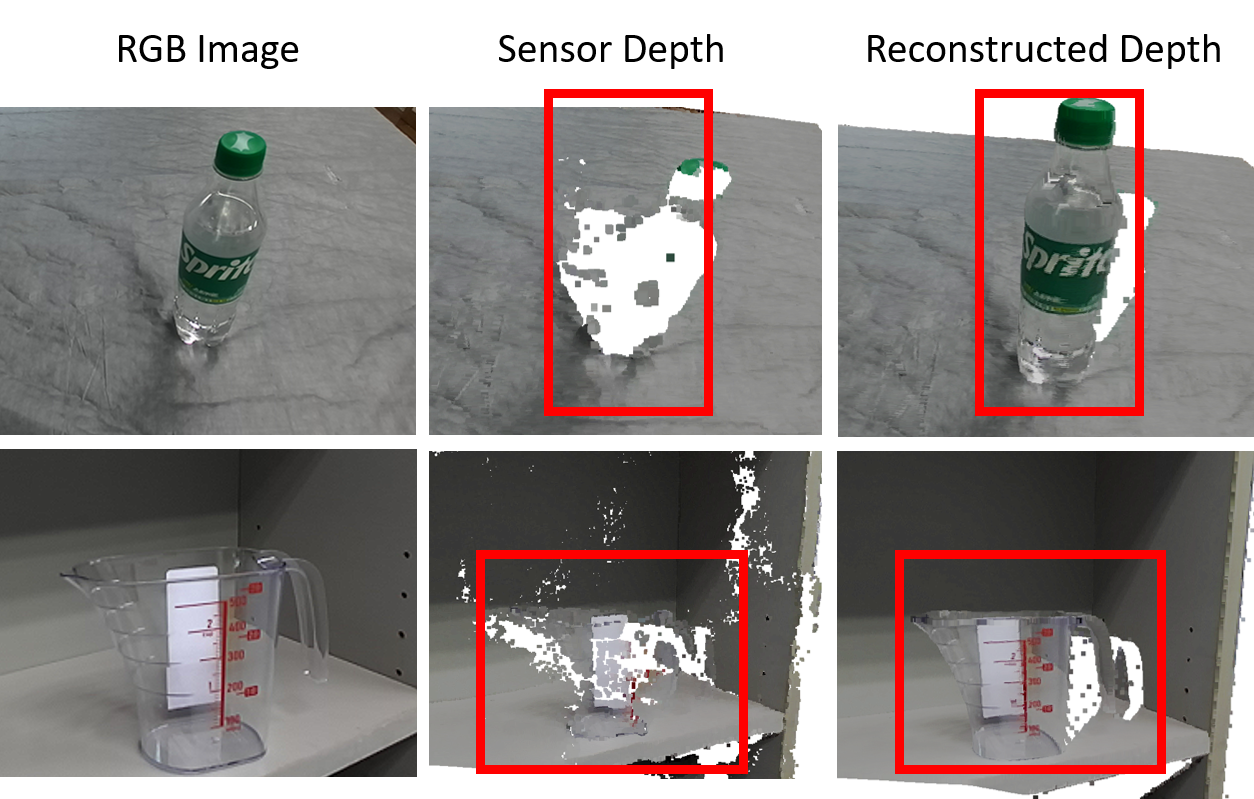}
\caption{Comparison of ZED camera's depth and reconstructed depth.} 
\label{fig:depth_comparison}
\vspace{-15pt}  
\end{figure}

\noindent{\textbf{Transparent Depth Estimation.}}  \label{subsubsec:depth_estimation}
Fig.~\ref{fig:depth_comparison} shows the challenge that raw sensor depth from ZED stereo camera~\cite{tadic2022perspectives} fails to capture the surface depth of transparent objects. To address this limitation, we harness FoundationStereo~\cite{wen2025foundationstereo, wen2025fast}, a foundation model for stereo depth estimation. 
It takes stereo images from the ZED camera as input and outputs pixel-wise metric-scale depth.
The reconstructed depth enhances the overall robot manipulation pipeline by improving object and hand pose estimation, as well as high-quality 3D collision map reconstruction for the motion planner.

\noindent{\textbf{Open-Vocabulary Segmentation.}}  We utilize open-vocabulary detection~\cite{liu2023grounding, khanam2024yolov11} to obtain bounding boxes of hands and objects from language descriptions, followed by a segmentation model~\cite{ravi2024sam} that generates detailed masks using these boxes as prompts for pose estimation.

\noindent{\textbf{Mesh Database.}} 
We pre-build object mesh databases containing textured object shapes to be used for object pose estimation during both demonstration video parsing and robot execution. Meshes are obtained via image-to-3D reconstruction~\cite{xu2024instantmesh, hunyuan3d2025hunyuan3d} and existing databases~\cite{kim2024transpose}. 
Each mesh is canonicalized and aligned to the coordinate frame of the estimated hand pose from the demonstration video, ensuring consistent trajectory retargeting across novel objects. This database contains geometry information with textures, serving exclusively for robot execution pose estimation.

\noindent{\textbf{Novel Object Pose Estimation and Tracking.}} 
A sequence of object poses represents the movement of an object.
For example, in a pouring task, the trajectory of the 6D object pose contains a complete pouring motion trajectory performed by a human. We use state-of-the-art novel object pose and tracking methods~\cite{wen2024foundationpose}. It takes as input the reconstructed depth, segmentation masks, and 3D meshes, which are obtained as aforementioned. 

\noindent{\textbf{Hand Pose Estimation.}} 
The purpose of hand pose estimation is to transfer object pose trajectories to robot action coordinates, as well as to compute grasp poses. We use a hand pose detector~\cite{potamias2025wilor} to extract 21 keypoints. We then adjust their scale using our reconstructed depth from stereo and rendered hand depth, given the estimated MANO hand mesh, to obtain accurate 3D hand joint positions.
From the rectified keypoints, we construct a wrist coordinate system using key hand landmarks. The z-axis is computed as the cross product of the thumb and index fingertip directions from the wrist. The y-axis represents the mean of these two directional vectors, capturing the middle orientation. The x-axis is calculated as the cross product of the y-axis and z-axis to ensure orthogonality. All vectors are normalized to unit length, and the hand translation is defined as the center point between the thumb and index fingertips.

\noindent{\textbf{Trajectory Database.}} 
Based on hand pose guidance, we transform object pose trajectories using action-specific reference frames: the target container object pose for pouring, the initial pose for pick, and the final pose for placement.
The transformed trajectories are then processed for storage. Since raw pose trajectories are dense and noisy, we subsample them at every 2cm or $5^{\circ}$ difference and apply smoothing to ensure stable robot manipulation. 
The processed trajectories are stored in the trajectory database. 
In total, we extracted three trajectories, each corresponding to the pick, place, and pour skill from single-object videos.

\subsection{Vision-Language Guided Task Planning} \label{subsec:task_plan}
The task planner takes a high-level task instruction as input and generates a task plan consisting of primitive actions (\eg Pick, Place, PlaceInFront, LookFor) for the robot to execute.
It consists of three main components: VLM planner, plan translator, and plan grounding  (Fig.~\ref{fig:overview} bottom-left). 
The VLM planner processes task instructions and visual inputs to produce an initial high-level plan. The plan translator converts this plan into the formal planning language PDDL \cite{fox2003pddl2}, \cite{helmert2009concise} while checking for syntactic errors. 
Finally, the plan grounding takes into account robot constraints (\eg limited FoV under eye-in-hand camera configuration), evaluates feasibility, and adds intermediate actions (\eg look for objects outside view) to ensure executability through iterative search-and-refinement.

\definecolor{blue}{RGB}{76, 115, 199}
\definecolor{orange}{RGB}{229, 124, 37}
\definecolor{red}{RGB}{244, 0, 0}
\begin{figure}[t]
\centering
\begin{subfigure}{\linewidth}
    \centering
    \includegraphics[width=0.95\linewidth]{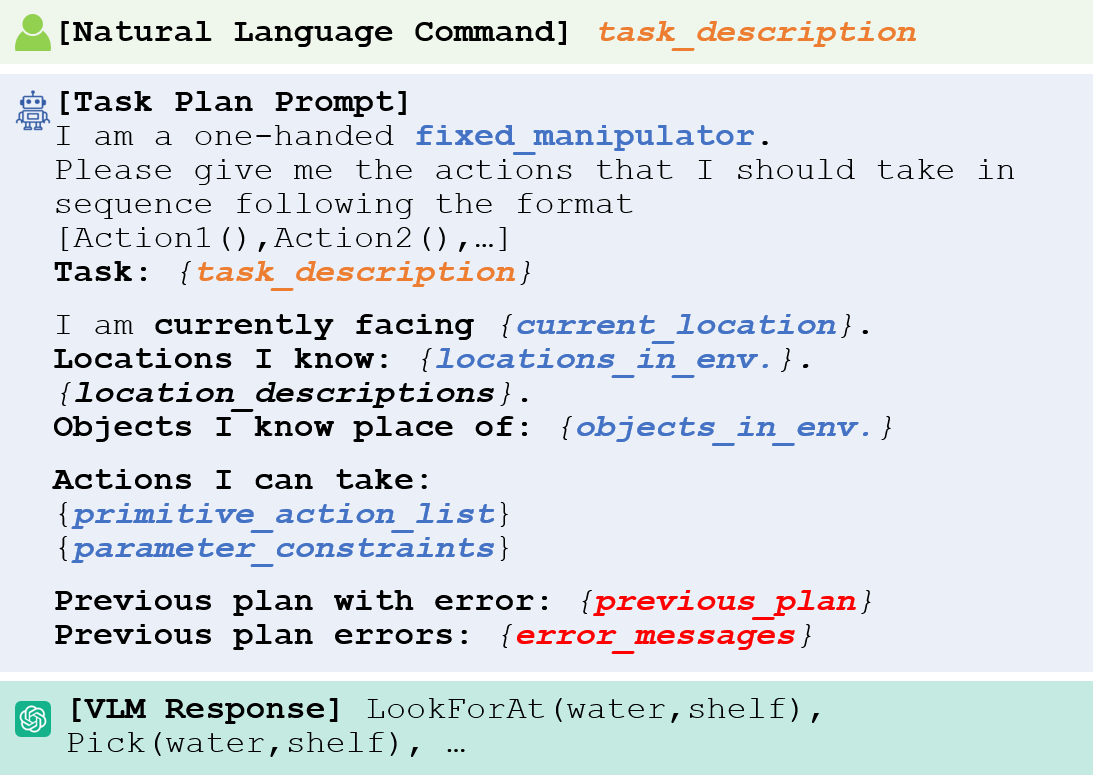}
    \vspace{-5pt}
    \caption{Prompt format for task planning and refinement. }
    \label{fig:prompt1}
\end{subfigure}
\begin{subfigure}{\linewidth}
    \centering
    \includegraphics[width=0.95\linewidth]{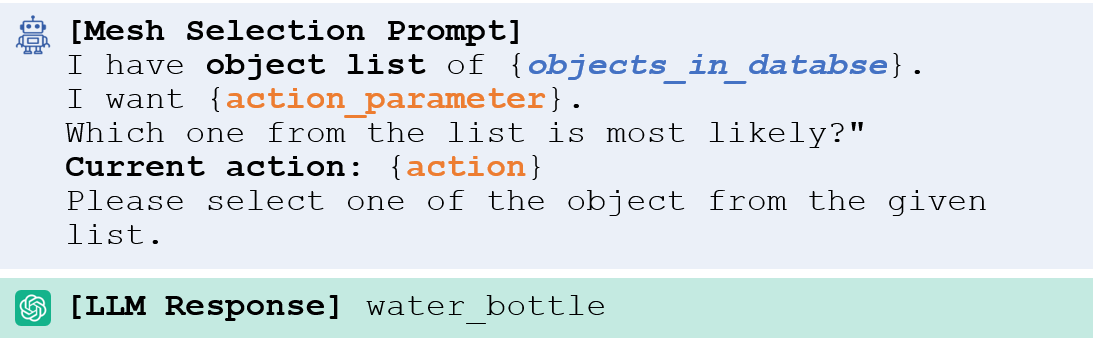}
    \vspace{-5pt}
    \caption{Prompt format for mesh data selection.}
    \label{fig:prompt2}
\end{subfigure}
\caption{VLM prompting process. \includegraphics[width=0.015\textwidth]{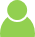} Human first provides the robot with a task description in natural language. \includegraphics[width=0.015\textwidth]{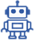} Robot then formulates a templated prompt and inquires \includegraphics[width=0.015\textwidth]{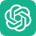} VLM for responses. \textcolor{blue}{Blue}: context information including robot state, primitive actions, and environment state. 
\textcolor{orange}{Orange}: task-dependent prompts. 
\textcolor{red}{Red}: error messages and invalid plans.}

\label{fig:prompt}
\vspace{-18pt}
\end{figure}

\noindent{\textbf{Task Plan Definition.}}
The task planning is formalized as $\langle s_{init}, A, E, I_{init}, D \rangle$, where the goal is to generate a plan $\pi = [a_1, \dots, a_T], \; a_t \in A$ that fulfills the task description $D$, starting from the initial state $s_{init}$ and utilizing the primitive action set $A$, environment information $E$, and the initial images $I_{init}$. 
The robot operates using primitive actions $A$ (obtained in Sec.~\ref{subsec:human_demonstration}), where each action produces a deterministic state transition.
The environment information $E$ includes task locations (e.g., staging area) and the pose of fixed objects (e.g., shelf), while manipulatable objects are inferred by the VLM from images and commands.
The robot state $s$ consists of (1) the current camera facing direction, (2) whether an object is held in the gripper, (3) a list of saved objects with their poses, and (4) the robot’s joint configuration.

\noindent{\textbf{VLM Planner.}} 
The VLM outputs a sequence of actions $\pi_{VLM}$, where each action consists of an action type (\eg \textit{Pick}) and a target object or location.
As shown in Fig.~\ref{fig:prompt1}, the prompt specifies the task goal, known locations, objects, and available actions, along with representation constraints and a fixed format to ensure a clear and ordered action plan. 
This sequence serves as a skeleton plan~\cite{yang2025guiding}, which is subsequently grounded and refined.
We utilize a foundation VLM~\cite{achiam2023gpt} without fine-tuning to preserve generalization and maintain robust reasoning across diverse tasks.

\noindent{\textbf{Plan Translation.}}
$\pi_{VLM}$ is converted into a list of actions represented in PDDL format~\cite{helmert2009concise} in plan translation.
Each primitive action consists of an action type, parameters $o$ (target objects and locations), a \textit{precondition} $PRE(a, s_t,o,E)$ that must be satisfied before execution, 
and an \textit{effect} $EFF(a, s_t,o)$ that represents the resulting state change. 
During this conversion, the translator verifies (1) whether the plan adheres to the defined primitive actions and (2) whether all required parameters for each action are satisfied.
If either condition is not met, the translator returns an error message and prompts the VLM to regenerate the plan.

\begin{algorithm}[th]
\caption{Grounded Plan Search}
\label{alg:iterative_backward_bfs}
\footnotesize
\begin{algorithmic}[1]
\State \textbf{In:} $\Pi$, $s_{init}$, $E$, $\textsc{MaxNodes}$ \quad \textbf{Out:} $\pi_g$
\State $\pi_g\gets\emptyset$
\For{$i=1\to N$}
  \State \textcolor{black}{$(A_k,A_c)\gets\Call{Classify}{\pi^i}$}; $\pi_c\gets\emptyset$
  \ForAll{$k\in A_k$}
    \State $\pi^\star\gets\pi_c\cup\{k\}$; $Q\gets(\pi^\star,\emptyset)$; $V\gets\emptyset$; $n\gets0$
    \While{$Q\neq\emptyset$}
      \State $(\pi,U)\gets\Call{Pop}{Q}$
      \State \textbf{if} $\pi\in V$ \textbf{then continue}; $V\gets V\cup\{\pi\}$
      \State \textcolor{black}{$(f,A_s)\gets\Call{CheckFeasible}{\pi,E,s_{init}}$}
      \State \textcolor{black}{\textbf{if} $f$ \textbf{then} $\pi_c\gets\pi$; \textbf{break}}
      \State \textcolor{black}{$Q\gets Q\cup\Call{UpdateQue}{\pi,A_c,A_s,U,V}$}
      \State \textcolor{black}{\textbf{if} $++n\ge\textsc{MaxNodes}$ \textbf{then return} $\pi_g$}
    \EndWhile
    \State \textcolor{black}{\textbf{if} $\pi_c=\emptyset$ \textbf{then return} $\pi_g$}
  \EndFor
  \State \textcolor{black}{$\pi_g\gets\pi_g\cup\pi_c$}
\EndFor
\State \textcolor{black}{\Return $\pi_g$}
\end{algorithmic} 
\end{algorithm} 

\noindent{\textbf{Grounded Plan Search.}}
While the VLM shows promising reasoning and generality, it still has two key limitations:  
(1) It often overlooks robot-specific constraints (e.g., one-handed or eye-in-hand camera systems) and (2) may omit steps required for execution (e.g., placing an object down before picking another).  
To address these limitations, we (1) evaluate the feasibility of $\pi_{VLM}$ and (2) search for missing actions to produce a complete, executable plan (e.g., \textit{Pick}(target object) $\rightarrow$ \textit{Place}(object in hand, place), \textit{LookFor}(target object), \textit{Pick}(target object)), as described in Algorithm~\ref{alg:iterative_backward_bfs}.
We ensure executability by sequentially validating each \textit{precondition} of action while updating the robot state via the corresponding \textit{effect}:
\begin{equation}
s_{t+1} = \textit{EFF}(a, s_t, o), \quad \text{valid if } \textit{PRE}(a, s_t, o, E) \subseteq s_t
\end{equation}

We divide $\pi_{VLM}$ into subtasks $\Pi=\{\pi^1,\dots,\pi^N\}$ at placement actions, ensuring free robot hands for independent subtask execution.  
Each subtask labels object manipulation actions (e.g., \textit{Pick}) as key actions $A_k$ and others (e.g., \textit{LookFor}) as connecting actions $A_c$.  
Key actions are validated sequentially; if an action fails, a backward breadth-first search is triggered to find satisfying predecessor sequences using $A_c$ and the order-independent primitive actions $A_s$.
This process includes actions to move the robot joints into suitable configurations when they are not appropriate for executing the next action.
While performing search, it tracks the visited plans $V$ and records used actions $U$.
Search failure returns unsatisfied preconditions and partial plans to VLM for replanning.

This plan grounding step enables the planner to recover from robot-constraint errors while keeping order of key actions. It also reduces the time cost caused by repeated VLM queries and mitigates overfitting, where the VLM gradually shifts focus from the original task goal to resolving robot constraints due to repeated interactions.

\noindent{\textbf{Plan Refinement}}
bridges VLM reasoning and robot execution through iterative feedback. 
Refinement occurs in two cases: (1) PDDL syntax errors during translation, generating format correction messages like "\textit{Failed to create \{action\} instance: \{error\}.}"; and (2) task planning errors during search when connecting actions cannot be found, prompting precondition error messages with partial plans. 
This process iterates until producing a fully grounded, executable plan, as shown Fig.~\ref{fig:overview}.

\begin{figure}[t]  
\centering  
\includegraphics[width=0.95\linewidth]{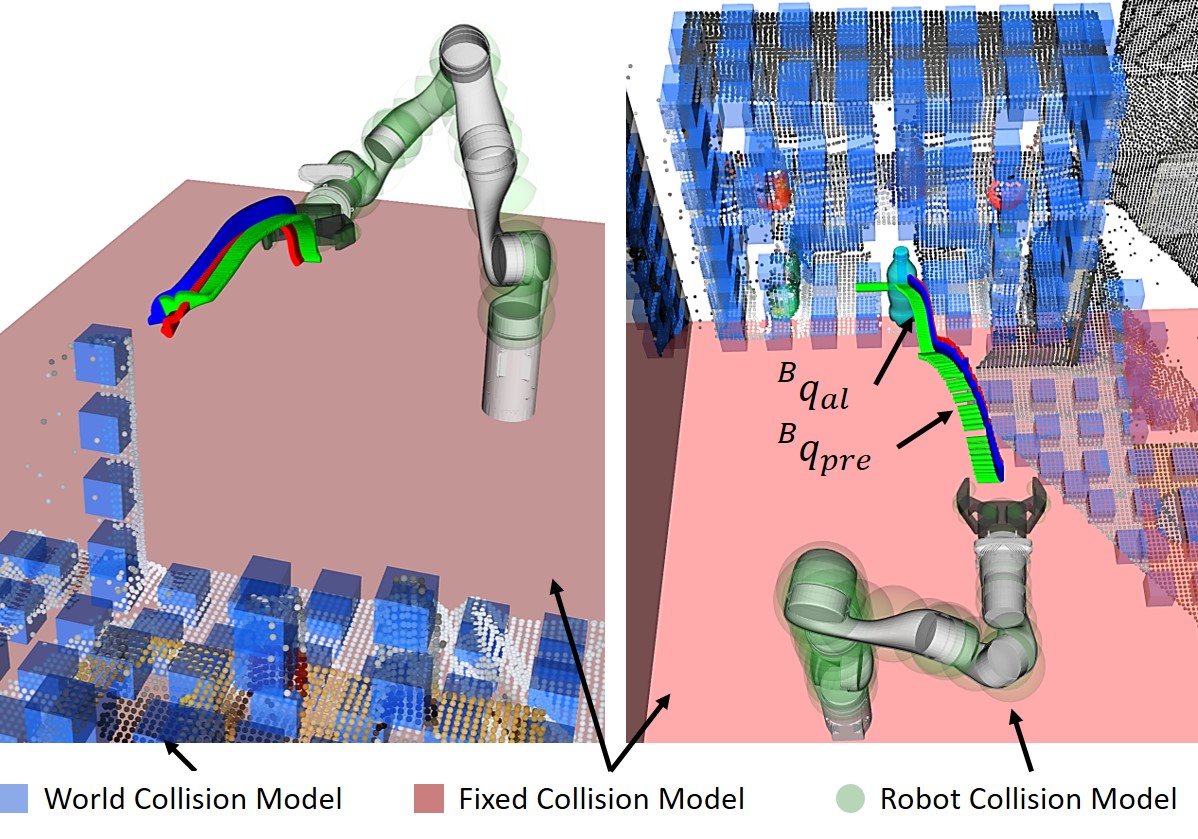}
\caption{Last-Inch Motion Planner: Pouring (Left) and Pick (Right). RGB axes visualize planned end-effector poses. Blue boxes represent the approximated collision map derived from the point cloud.}
\label{fig:trajectory_change}
\vspace{-15pt}
\end{figure}

\subsection{Demonstration-Guided Robot Action Execution} \label{subsec:robot_execution}
The robot action execution stage implements the action sequence generated in Section~\ref{subsec:task_plan} through the trajectory and mesh database from Section~\ref{subsec:human_demonstration}. This execution framework integrates three core components: (1) target object pose estimation, (2) target trajectory generation, and (3) collision-aware motion planning.

\noindent{\textbf{Action Manager}}
sequentially processes planner-generated actions by selecting target meshes and primitive skill, generating motions, and evaluating feasibility and termination. Target mesh selection maps VLM action parameters (\eg cola\_on\_the\_shelf) to corresponding mesh database objects (\eg cola) using an LLM with simple prompts, as shown in ~Fig.~\ref{fig:prompt2}. Action feasibility is evaluated based on predefined \textit{precondition} that depend on the current robot and environment states. Upon action completion, the action manager notifies the task planner to dispatch the next action.

\noindent{\textbf{Novel Objct Pose Estimation.}}
After retrieving the mesh through the action manager and the mesh 
database, we estimate the 6D object pose ${}^{B}\mathbf{T}^{obj}$ 
from reconstructed depth and open vocabulary segmentation, 
as in Sec.~\ref{subsec:human_demonstration}.

\noindent{\textbf{Demonstration-Guided Trajectory Generation.}}
Rather than simply replicating existing trajectories, our method retargets to different objects and adapts to varying environments through pose-based adaptation.
First, we generate an initial Cartesian-space trajectory ${}^{B}\tau_{\mathrm{init}}$ by mapping the demonstration trajectory ${}^{obj}\tau^{EE}$ from the object frame ($obj$) to the base link frame ($B$) using the estimated object pose, enabling reuse for novel objects with diverse poses. 
\begin{align}
    {}^{B}\tau_{\mathrm{init}} &= \big\{{}^{B}\mathbf{x}^{EE}_{t} = {}^{B}\mathbf{T}^{obj} \, {}^{obj}\mathbf{x}^{EE}_{t} 
          \;\big|\; {}^{obj}\mathbf{x}^{EE}_{t} \in {}^{obj}\tau^{EE} \big\}_{t=0}^{T}, \nonumber \\ 
    {}^{B}\mathbf{x}^{EE}_{t} &\in SE(3) 
\end{align}

Next, we apply a rotation-based alignment that keeps the final target position fixed at the manipulation pose of object while rotating the entire waypoint set, so that the starting pose of the trajectory aligns with the current end-effector pose of robot.
To compute this alignment, let $\mathbf{x}_\text{start}$, $\mathbf{x}_\text{target}$, and $\mathbf{x}_\text{cur}$ denote the original first waypoint, the final target waypoint, and the current end-effector position, respectively. We form two unit direction vectors:
\begin{equation}
    \mathbf{v}_{\text{orig}}=\frac{\mathbf{x}_\text{target}-\mathbf{x}_\text{start}}{\left\|\mathbf{x}_\text{target}-\mathbf{x}_\text{start}\right\|},\qquad
    \mathbf{v}_{\text{cur}}=\frac{\mathbf{x}_\text{target}-\mathbf{x}_\text{cur}}{\left\|\mathbf{x}_\text{target}-\mathbf{x}_\text{cur}\right\|}.
\end{equation}

Using rotation axis $\mathbf{v}=\mathbf{v}_{\text{orig}}\times \mathbf{v}_{\text{cur}},\quad$ and angle $c=\mathbf{v}_{\text{orig}}\cdot \mathbf{v}_{\text{cur}}$ with  Rodrigues' formula with $[\mathbf{v}]_\times$ the skew-symmetric matrix of $\mathbf{v}$, the rotation matrix is
\begin{equation}
    R = I + [\mathbf{v}]_\times + [\mathbf{v}]_\times^2\cdot\frac{1-c}{\|\mathbf{v}\|^2}.
\end{equation}
Each waypoint $\mathbf{x}_i$ is rotated about the fixed target:
\begin{equation}
    \mathbf{x}_i' = R\big(\mathbf{x}_i-\mathbf{x}_\text{target}\big)+\mathbf{x}_\text{target}, 
    {}^{B}\tau_{\mathrm{al}} = \{\mathbf{x}_i' \mid \mathbf{x}_i \in {}^{B}\tau_{\mathrm{init}}\}.
\end{equation}
This realigns ${}^{B}\tau_{\mathrm{al}}$ to the robot’s current configuration, enabling smooth execution without reparameterization.

\noindent{\textbf{Last-Inch Motion Planner.}}
The motion execution consists of two stages: (1) global planning to reach the start point of the demonstration-based trajectory and (2) following the last-inch demonstration-based trajectory. A point cloud is generated from the depth estimation results described in \ref{subsubsec:depth_estimation} and used to construct a world collision model for safe execution. This process is performed during the \textit{LookFor} action, when the robot has the best view of the environment.

In the first stage, a collision-aware joint-space trajectory ${}^{B}q_{\mathrm{pre}}$ is planned \cite{sundaralingam2023curobo}. 
This trajectory moves the robot from its current pose to the starting pose of ${}^{B}\tau_{\mathrm{al}}$. 
The final joint configuration from ${}^{B}q_{\mathrm{pre}}$ is then used as the initial condition for tracking ${}^{B}\tau_{\mathrm{al}}$. 
A corresponding joint-space trajectory ${}^{B}q_{\mathrm{al}}$ for ${}^{B}\tau_{\mathrm{al}}$ is generated using inverse kinematics (IK) formulated as a quadratic programming (QP) optimization problem, incorporating collision avoidance and joint limit constraints \cite{ashkanazy2023collision}. 
If a valid plan cannot be generated, ${}^{B}\tau_{\mathrm{al}}$ is adjusted by introducing small translation and orientation perturbations to the target object pose, 
to ensure a collision-free path for both ${}^{B}q_{\mathrm{pre}}$ and ${}^{B}q_{\mathrm{al}}$ as shown in Fig.~\ref{fig:trajectory_change}.

During the second stage, the robot tracks ${}^{B}\tau_{\mathrm{al}}$ using the same QP-based IK solver with adaptive accuracy.
The solver tolerance is gradually tightened as the end-effector approaches the target, prioritizing collision avoidance and joint limit handling during the early phase, and enforcing higher precision near task completion.
This strategy enables safe obstacle avoidance during the approach while ensuring accurate manipulation near task completion~\cite{lu2023cfvs}.


\begin{figure*}[t]  
\centering  
\includegraphics[width=0.95\linewidth]{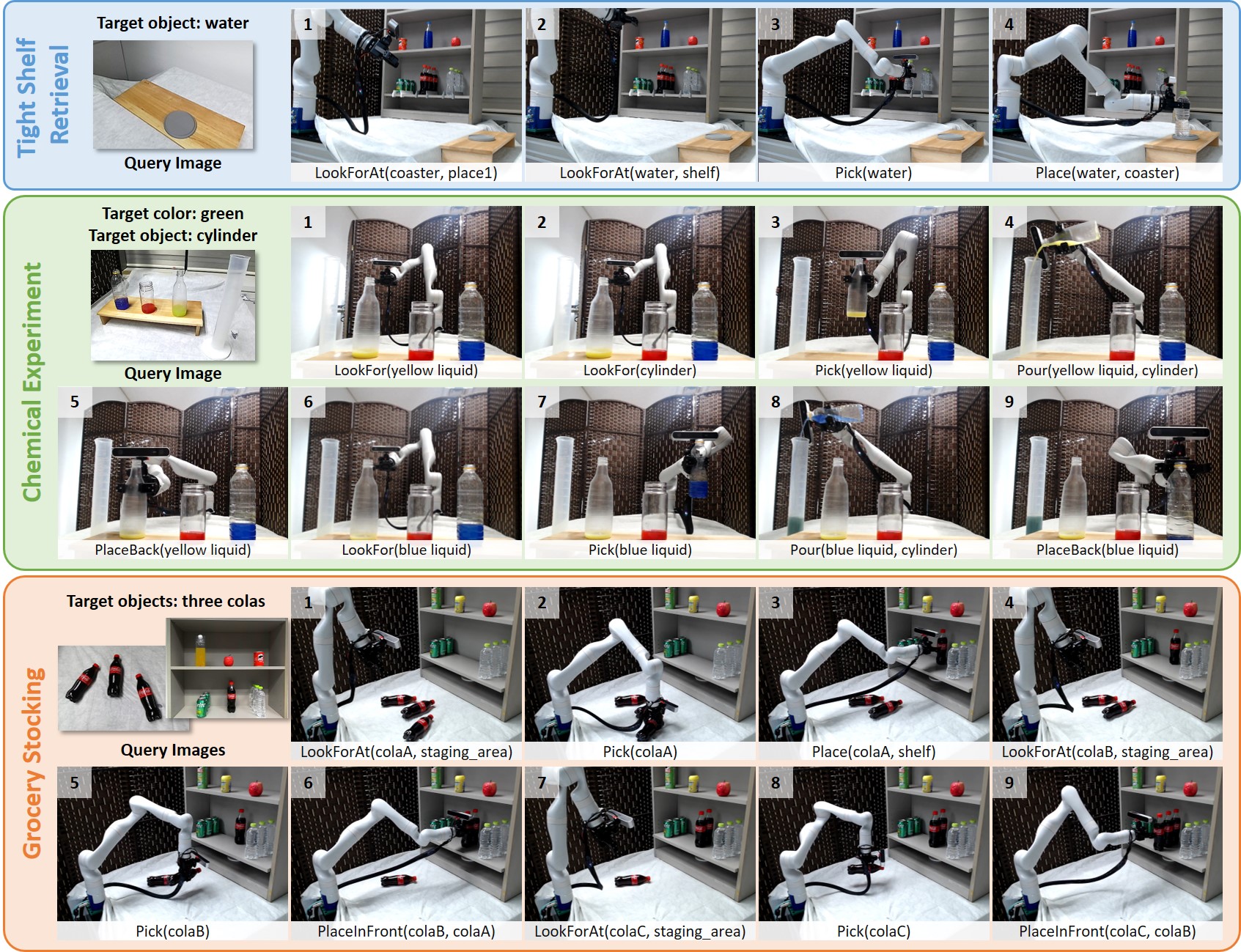}
\caption{Sequences of our three manipulation tasks with input query images and target objects in real-world environments.}
\label{fig:real_world_experiments}
\vspace{-15pt}  
\end{figure*}

\begin{figure}[t]  
\centering  
\includegraphics[width=0.8\linewidth]{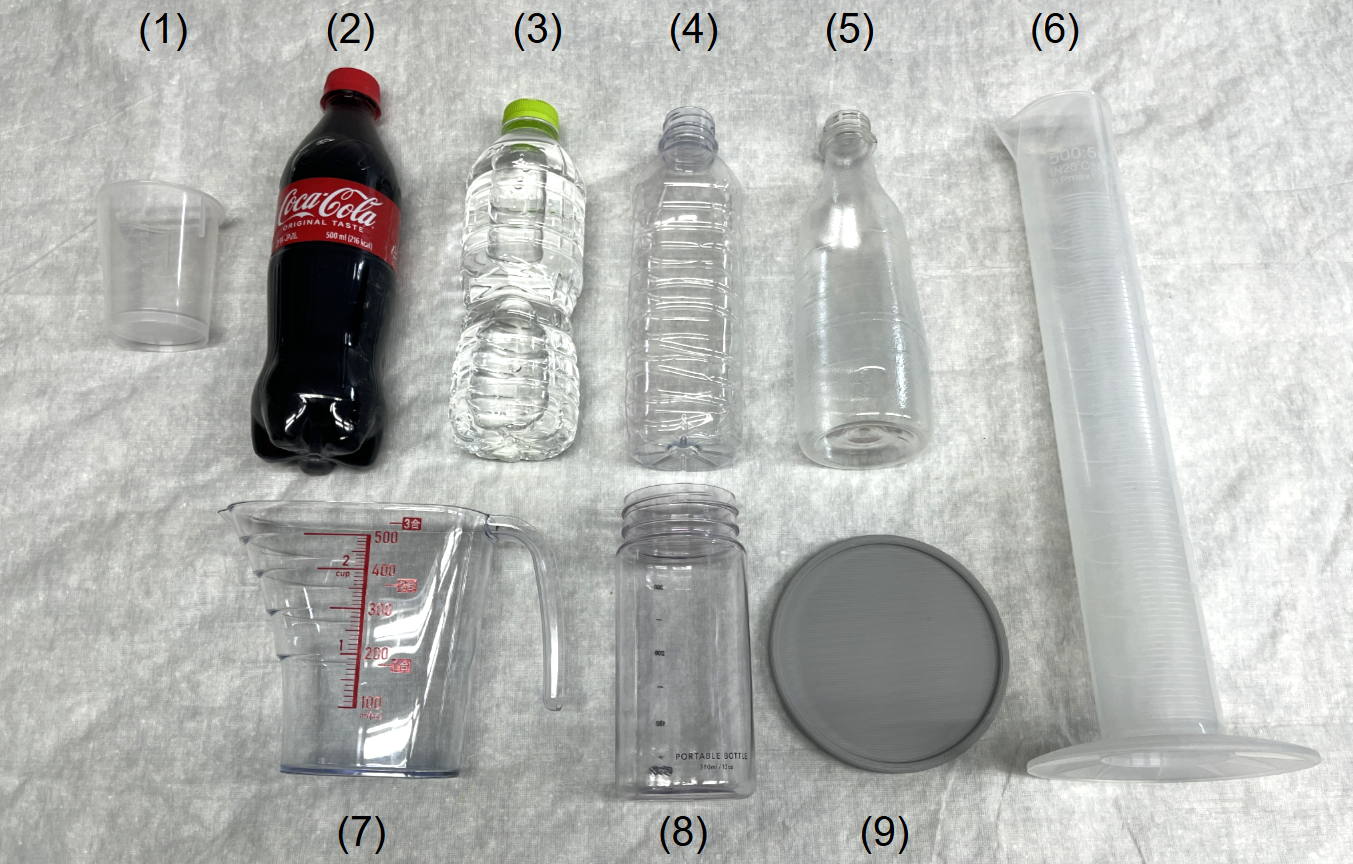}
\caption{Real-world objects used for manipulation experiments.}
\label{fig:dataset}
\vspace{-5pt}  
\end{figure}

\section{Experiments} \label{sec:experiments}

The evaluation is conducted through real-world experiments using a Kinova Gen3 7-DOF arm equipped with a Robotis RH-P12-RN gripper and an eye-in-hand ZED stereo camera. 
For evaluation, we consider three sequential manipulation tasks (Fig.~\ref{fig:real_world_experiments}) using 9 different transparent and non-transparent objects (Fig.~\ref{fig:dataset}).

\noindent{\textbf{Task1: Tight Shelf Retrieval.}} 
\textit{"Place the [target object] from the shelf onto the coaster."  }
This task requires precise pose estimation for accurate manipulation within tight shelf constraints. We evaluated three different transparent objects, focusing on short-horizon manipulation.

\noindent{\textbf{Task2: Chemical Experiment.}}
\textit{"Can you make [target color] liquid in the [target object]?"  }
This task simulates laboratory scenarios using transparent objects filled with various liquids. We tested with seven transparent objects and two target colors, requiring long-horizon planning, precise pouring, and collision avoidance in dense environments.
\noindent{\textbf{Task3: Grocery Stocking.}}
\textit{"Arrange the [objects] on the staging area to the shelf in a straight row as shown in the reference image, placing each one directly in front of the previous one." } 
Inspired by retail scenarios, this automatic organization task requires long-horizon planning, collision avoidance, and precise pose estimation for aligned stocking.

For each task, we conducted 10 trials with different object instances and pose variations. 
The three single-object trajectories (pick, place, pour) were extracted from the object shown in Fig.~\ref{fig:dataset}-(1).
Our framework supports 10 primitive actions (\textit{LookForAt}, \textit{LookFor}, \textit{Pick}, \textit{Pour}, \textit{PlaceBack}, \textit{Place}, \textit{PlaceBetween}, \textit{PlaceInFront}, \textit{Face}, \textit{InitPose}).
The environments include a shelf (\textit{Task 1}), a laboratory area (\textit{Task 2}), and a staging area with a shelf (\textit{Task 3}).

\subsection{Baselines}

We evaluated our approach against two baseline methods: ClearGrasp~\cite{sajjan2020clear} which performs transparent object manipulation through depth estimation, and YODO~\cite{wen2022you} which leverages human demonstrations with category-level object pose estimation. While both methods focus on short-horizon tasks, we equip them with our task planner and motion planner to enable long-horizon manipulation.
To ensure fair evaluation, we made necessary adaptations to both baselines. For ClearGrasp, which lacks pose estimation capabilities for precise manipulation, we integrated our pose estimation module and collision checking components. For YODO, it originally uses category-level pose estimation designed for industrial objects. We thus augment it with the state-of-art category-level pose estimation for transparent objects~\cite{chen2022stereopose}.

\subsection{Real World Results}

\begin{figure}[t]  
\centering  
\includegraphics[width=1.0\linewidth]{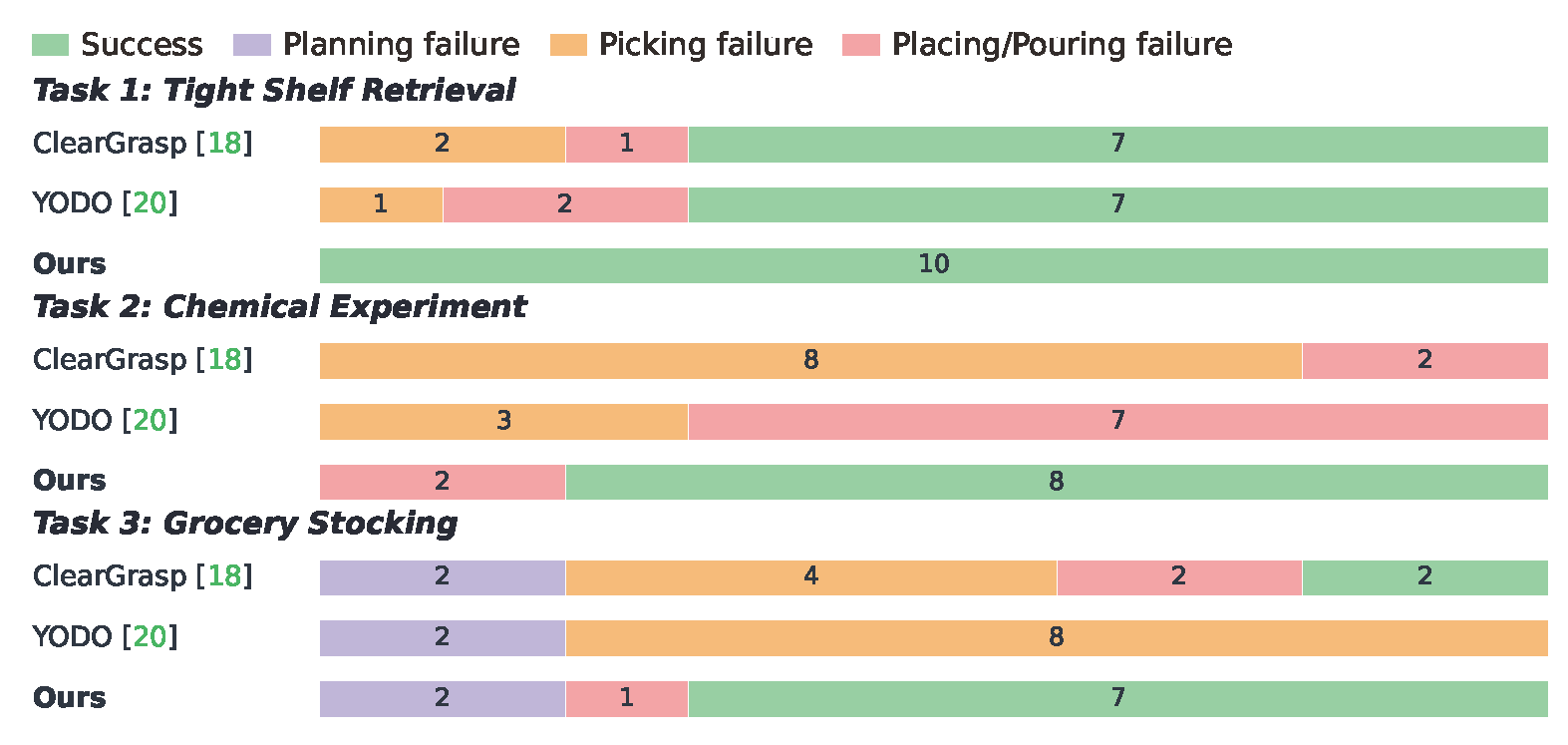}
\vspace{-13pt}
\caption{Comparison with state-of-the-art methods.}
\label{fig:sota_comparison}
\vspace{-10pt}  
\end{figure}

Fig.~\ref{fig:sota_comparison} summarizes the task success rates and failure breakdowns for each task. We observed that all methods perform well on Tight Shelf Retrieval (Task 1), since most existing methods focus primarily on this type of task. However, for long-horizon tasks involving precise manipulation—such as chemical experiment and grocery stocking—the performance of YODO~\cite{wen2022you} and ClearGrasp~\cite{sajjan2020clear} drops significantly.

Specifically, ClearGrasp~\cite{sajjan2020clear} relies on neighboring depth information to complete transparent regions, but when surrounding depth measurements are spatially distant, the resulting depth estimates become excessively noisy, hindering precise manipulation capabilities. Consequently, the majority of pouring tasks failed, with limited success observed only in grocery stocking scenarios where transparent objects have nearby reference surfaces.

YODO~\cite{wen2022you} fails in chemical experiment and grocery stocking tasks mainly because category-level object pose estimation does not generalize well to in the wild real-world scenarios without fine-tuning. Additionally, some irregular transparent objects cannot be easily categorized into predefined category names. For such objects, we performed pose estimation using the closest matching category. 
This highlights the limited generalization of category-level pose estimation methods and weak adaptiveness to ambiguous object types, which occur frequently for transparent object manipulation.
Furthermore, the lack of collision checking capabilities degrades performance, as collision avoidance is essential for safe manipulation in cluttered environments.

\subsection{Ablation Studies} \label{subsec:ablation_study}
In our ablation studies, we address the following key questions for transparent robotics manipulation:    

\noindent{\textbf{(1) Is depth estimation necessary for transparent object manipulation?}} To answer this question, we evaluate our method using raw depth sensor observations instead of our depth estimation results (Fig.~\ref{fig:depth_comparison}). The comparison between Table~\ref{tab:ablation_study}-(1) and Table~\ref{tab:ablation_study}-Ours demonstrates that our depth estimation significantly improves performance across all tasks. 
Specifically, in the grocery stocking task, inaccurate depth information hindered the robot from compactly arranging objects within the shelf dimensions. This verifies the significant benefits of enhanced depth estimation for precise transparent robotics manipulation, as raw depth sensors alone are insufficient for handling transparent objects.

\begin{table}[t]
\centering
\caption{Ablation study of our core components on three different tasks in the real world.}
\setlength{\tabcolsep}{4pt} 
\renewcommand{\arraystretch}{1.1} 
\resizebox{\linewidth}{!}{ 
\begin{tabular}{l|cccc|ccc}
\toprule
Method & \multicolumn{4}{c|}{Core components} & \multicolumn{3}{c}{Success Rate (\%) ↑} \\
& Depth & 6D Pose & \makecell{Plan\\Refinement} & \makecell{Plan\\Search} & Task 1 & Task 2 & Task 3 \\
\hline
(1) &  & \greencheckmark & \greencheckmark & \greencheckmark & 60 & 30 & 20 \\
(2) & \greencheckmark &  & \greencheckmark & \greencheckmark & 70 & 60 & 0 \\
(3) & \greencheckmark & \greencheckmark &  &  & 0 & 10 & 0 \\
(4) & \greencheckmark & \greencheckmark & \greencheckmark & & 100 & 10 & 10 \\
\rowcolor[gray]{0.9} \textbf{Ours} & \greencheckmark & \greencheckmark & \greencheckmark & \greencheckmark & \textbf{100} & \textbf{80} & \textbf{70} \\
\toprule
\end{tabular}}
\label{tab:ablation_study}
\vspace{-15pt}
\end{table}
 
\noindent{\textbf{(2) Is 6D pose estimation necessary compared to 3D position-based manipulation?}} One of the straightforward approaches for robot manipulation uses 3D position from open vocabulary segmentation and projects to 3D using depth for manipulation. As shown in Table~\ref{tab:ablation_study}-(2), this position-based method achieves reasonable performance in Tight Shelf Retrieval (70\%) and Chemical Experiment (60\%) tasks, but fails in grocery stocking tasks, demonstrating the limitations of position-only approaches for grasping diverse rotation. 
Specifically, when objects are lying with small rotation errors, grasps often fail, highlighting the necessity of accurate 6D pose for robust manipulation.

\noindent{\textbf{(3-4) Can VLMs effectively handle long-horizon task planning?}} 
While naive VLMs may appear capable of long-horizon task planning, LLMs lack the ability to incorporate robot manipulation constraints into action sequences
For example, in a pick-and-place task, the model may attempt to place an object before actually grasping it. 
Table~\ref{tab:ablation_study}-(3) shows that a naive LLM mostly fails to generate valid plans, even for a short-horizon task such as tight shelf retrieval. 
An alternative approach is to iteratively refine the plan using a validator and repeated queries with error feedback, as in prior studies~\cite{zhou2024isr}.
However, Table~\ref{tab:ablation_study}-(4) demonstrates that this approach still struggles with long-horizon tasks, such as chemical experiments and grocery stocking, when the number of refinement iterations is limited to 10. 
Even when task planning succeeds, execution failures can still occur during the motion planning stage because this method relies solely on symbolic states without incorporating continuous variables such as joint poses, unlike our search-based method. 
Compared to Table~\ref{tab:ablation_study}-(3-4) of VLM planning, our proposed search-based refinement successfully handles long-horizon planning.

\section{Conclusion}
We propose DeLTa, a Demonstration and Language-Guided Novel Transparent Object Manipulation framework, that integrates novel object 6D pose estimation and precise long-horizon manipulation of transparent objects from language instructions. 
Our last-inch motion planner generalizes 6D trajectories to novel objects from single-object demonstrations, while our VLM-guided planner grounds task plans in robot configurations and refines them for long-horizon manipulation tasks.
Our method demonstrates robust transparent-object manipulation capabilities in real-world environments and challenging tasks, substantially outperforming existing competitive methods.

\noindent{\textbf{Limitations:}} First, DeLTa is currently limited to rigid object manipulation, as our pose estimation method assumes a rigid-body model. 
Second, our current implementation relies on pre-built canonicalized meshes for cross-object trajectory transfer and supports three primitive skills and ten primitive actions as a proof of concept. Extending both the mesh preparation pipeline and task diversity would be promising directions for future work.

{
    \scriptsize
    \bibliographystyle{IEEEtran}
    \bibliography{reference}
}

\end{document}